\documentclass[journal]{IEEEtran}

\usepackage{graphicx}
\usepackage[caption=true]{subfig}
\usepackage{pdflscape}
\usepackage{makecell}
\usepackage{multirow}
\usepackage{hhline}
\usepackage{amsmath}

\usepackage[hidelinks]{hyperref}
\newcommand{\mat}[1]{\mathbf{#1}}

\begin{document}

\author{Bernhard~Gahr$^{*}$,
        Shu~Liu$^{+}$,
        Kevin~Koch$^{*}$,
        Filipe~Barata$^{+}$,
        Andr\'{e}~Dahlinger$^{*}$,
        Benjamin~Ryder$^{+}$,
        Elgar~Fleisch$^{+}$,
        and~Felix~Wortmann$^{*}$% <-this % stops a space
\thanks{$^{*}$Institute of Technology Management, University St. Gallen, Switzerland \{bernhard.gahr, kevin.koch, andre.dahlinger, felix.wortmann\}@unisg.ch}%
\thanks{$^{+}$Department of Management, Technology and Economics, ETH Zurich, Switzerland \{liush, fbarata, bryder, efleisch\}@ethz.ch}
}
\markboth{}%
{Shell \MakeLowercase{\textit{et al.}}: Bare Demo of IEEEtran.cls for IEEE Journals}

\title{Driver Identification via the Steering Wheel}

% If necessary modify the number of words per table or figure default is set to
% 250 words per table and figure
% \setcounter{wordspertable}{250}
% \setcounter{wordsperfigure}{250}

% If words are counted manually, put that number here. This does not include
% figures and tables. This can also be used to avoid problems with texcount
% program i.e. if one does not have it installed.
% \setcounter{textwords}{200}

\maketitle

\begin{abstract}
Driver identification has emerged as a vital research field, where both practitioners and researchers investigate the potential of driver identification to enable a personalized driving experience. % as well as associated privacy issues.
Within recent years, a selection of studies have reported that individuals could be perfectly identified based on their driving behavior under controlled conditions. %, i.e., data collection on the basis of predefined driving tracks.
However, research investigating the potential of driver identification under naturalistic conditions claim accuracies only marginally higher than random guess.
The paper at hand provides a comprehensive summary of the recent work, highlighting the main discrepancies in the design of the machine learning approaches, primarily the window length parameter that was considered.
Key findings further indicate that the longitudinal vehicle control information %(throttle \& brake pedal) 
is particularly useful for driver identification, leaving the research gap on the extent to which the lateral vehicle control %(steering wheel) 
can be used for reliable identification.
%Building upon existing work, we provide evidence that reliable driver identification can be achieved even in naturalistic field settings with data limited to the steering wheel only.
%A novel approach for the design of the window length parameter is presented and assessed based on analyses of the steering wheel signal.
Building upon existing work, we provide a novel approach for the design of the window length parameter that provides evidence that reliable driver identification can be achieved with data limited to the steering wheel only.
The results and insights in this paper are based on data collected from the largest naturalistic driving study conducted in this field.
Overall, a neural network based on GRUs was found to provide better identification performance than traditional methods, increasing the prediction accuracy from under 15\% to over 65\% for 15 drivers.
When leveraging the full field study dataset, comprising 72 drivers, the accuracy of identification prediction of the approach improved a random guess approach by a factor of 25.

\end{abstract}

\section{Introduction}
%Motivation / Goal
%Driver identification is considered a complimentary step towards complex classifications of vehicle operation behavior.
Driver identification is considered as the first step towards analyzing driver behavior, including the realm of complex classifications of vehicle operation behavior.
For example, detecting driver distraction or fatigue for safety enhancement features~\cite{Iranmanesh2018}, or enabling systems that identify critical health issues such as heart or severe hypoglycemia attack.
The frictionless identification of individual drivers enables various benefits, including a streamlined driving experience without the need for cumbersome authentication, improved security through intrusion detection systems, and personalized in-car service models~\cite{Wurster2015}.
Naturally, automated identification raises questions regarding privacy --- especially due the increasing interconnectivity of modern cars~\cite{Enev2016} --- and the recent work of multiple research groups continues to show growing interest in the potential of driver identification.

The high potential of utilizing CAN-bus data collected from the vehicle has been the main focus of recent research on driver identification~\cite{Hallac2016, Hallac2018}.
For example, specialized analysis methods for a single CAN-bus signal can significantly improve identification accuracy~\cite{Gahr2018a, Marchegiani2018}, particularly when considering the brake pedal and throttle signal.
Some extreme results have claimed the ability to perfectly identify a driver~\cite{Ezzini2018, Wang2017} despite using just one CAN-bus signal such as the brake pedal~\cite{Enev2016}.
These results indicate the power of driver identification via longitudinal vehicle control.
However, all studies on this topic come with a variety of limitations.
These include study design issues, typically from collecting data under a controlled driving setting.
Research based on naturalistic driving data has not been able to recreate the extreme results.
Further, current research based on naturalistic driving data and relying on a multitude of sensors suffers from a potential misfitting of the trained model onto the car and not the driver~\cite{Wang2017}.
Finally, research limited to a single sensor suffers from practical issues, such as the necessity to observe the driver for an unsuitably long time until reliable identification can be made~\cite{Gahr2018a}.
Additionally, even low- and mid-range cars, such as the Volkswagen Golf or the Ford Fusion, currently come equipped with adaptive cruise control that takes over the longitudinal vehicle control.
Therefore, signals from the brake and gas pedals will no longer reveal any strong information about the driver, rendering existing identification models and research obsolete.
In order to advance driver identification and remain relevant in the coming years, new classification methods and strategies must be considered.
Specifically, those signals whose inputs are likely to remain in the hands of the driver, i.e. lateral control of the vehicle from the steering wheel, are set to become the only reliable information sources for driver identification.

In this paper we introduce an approach that leverages the steering wheel for driver identification.
Our research builds upon on an extensive literature review, which, to our best knowledge, does not yet exist for the topic of driver identification.
We show that the existing methods utilized in previous work, typically assessed in controlled driving settings, could not deliver their reported accuracies when applied to real-world naturalistic driving data.
Against this background, we present and evaluate a machine learning approach that performs well in this challenging setting, and which is motivated and based on statistical analyses of the steering wheel signal.

%Summary
This paper is structured as follows: we initially present research related to the topic of personal identification via vehicle data and highlight the existing research gap.
We then describe the study design and field test setting, followed by the analytical methodologies we applied.
Further, we present results validating the reviewed methods under a naturalistic setting, then show how identification accuracy improves when the presented novel feature collection method is applied.
Finally, we conclude with a discussion of the results and the implications of this research.

\section{Related Work}
\label{rel_work}
This section provides an extensive literature review.
We extended this section compared to other publications, because the topic of driver identification using CAN-bus data has become almost a trending topic in recent years.
While the first research was done in 2005~\cite{Wakita2005}, sixteen out of the twenty publications we found appeared in the last four years.
However --- to the best of our knowledge --- there is no comprehensive literature research, leading to partially overlapping and partially contradictory published research.
We first present research based on simulator and controlled field study data.
In Section~\ref{nat_studies} we present research based on naturalistic driving data.
A compact overview of the research is presented in Table~\ref{rel_work_tab} in the Appendix.
For the sake of compactness we neglect a closer description of works that used sitting posture~\cite{Riener2008}, finger vein~\cite{Wu2009a} or voice~\cite{Wu2009b} signals to identify a driver and only describe research that relied on data available on the CAN-bus.

\subsection{Simulator- and Controlled Driving Studies}
\label{sim_studies}
%We highlight hereby the test environment by simulator data, controlled driving data, i.e. CAN-bus data from a real car but collected with specific driving instructions for the study participants, and naturalistic driving data, i.e. real world CAN-bus data without any driving instructions for the study participants.
The first research on driver identification based on CAN-bus signals collected through a driving simulator led to the identification of drivers with an accuracy of up to 73\% for 30 drivers~\cite{Wakita2005}.
To achieve this result, the authors consider driving behavior signals including the accelerator pedal, brake pedal, and vehicle velocity.
Additionally they took following distance into account.
Later, the same group was able to optimize re-identification to 89.6\% for the simulator, which dropped to 71\% for real car data, by applying spectral analysis methods~\cite{Miyajima2006}.
In a study of 276 drivers measuring simulator- and real-car signals as well as the following distance, the group was able to re-identify a driver with 76.8\% accuracy~\cite{Miyajima2007}.

Almost ten years later, the next study focused on the implementation of neural nets of driver identifying methods~\cite{Martinez2014}.
The authors used data from a controlled field study~\cite{UYANIK2008}, where drivers had to follow a predefined route of 25~km.
Results were limited to detecting subsets of three to five of the available eleven drivers, reaching identification accuracies from 70\% to 85\% for different settings.
In a second step~\cite{Martinez2016} they used an extreme learning machine (ELM) network which was applied on \mbox{audio-}, \mbox{video-}, inertial measurement \mbox{unit-}, frontal laser \mbox{scanner-}, and CAN-bus signals.
Identification accuracies increased for two to eleven drivers to a range of 85\% to 97\%.
In the same year the first research group claimed perfect identification for a set of 15 drivers~\cite{Enev2016}.
The claim was made even stronger by achieving perfect identification using only 15 minutes of driving data, or 90 minutes of driving data from the brake pedal only.
Using only the steering wheel signal, the group achieved an accuracy of 83.33\% for 90 minutes of training data.
As in the previous research, the data was collected in a controlled field study.
These results are often considered state-of-the-art, since many later publications based their work on them.
In~\cite{Burton2016}, a second group claimed perfect identification not only after each session, but by continuously authenticating a driver throughout a driving session.
Their results were based on data from a simulator study.
Similar to~\cite{Burton2016}, \cite{Ezzini2018} focused on the immediate identification of a driver by using as few data as possible and claimed perfect identification of six to ten drivers using only five minutes of training data from controlled driving studies~\cite{kwak2016, Schneegass2013, romera2016}.

In~\cite{Jafarnejad2017} the researchers also focused on the immediate identification of a driver.
However, similar to~\cite{Enev2016} they contributed with analysis on a sliding window design for a random forest (RF) algorithm.
In contrast to the optimal 3-second window in~\cite{Enev2016}, they found their optimal window length at around 15 seconds, with an accuracy of 89\% for 15 and 82\% for 35 drivers.
In~\cite{Jeong2018} the authors focused again on the real-time identification of a driver using convolutional neural networks (CNN).
They argued that performances for larger sets drop using conventional RF, support vector machine (SVM) or ELM algorithms.
With 75 seconds of training data and a 0.25 seconds sliding window design they achieve 88\% accuracy for four drivers.
In~\cite{Marchegiani2018} the researchers focus on the design of SVMs and universal background models (UBM) to identify four drivers using only brake and throttle signals from a controlled driving study.
They achieve an identification rate of 83\% by taking an approximate window length of 5 seconds for the throttle and 15 seconds for the brake.
In~\cite{Rettore2018} the researchers aim to build an intrusion detection system based on CAN-bus values.
For 10 drivers, recognition accuracies over 98\% were achieved using extra trees (ET).
In~\cite{Bernardi2018} the authors introduce a multi-layer perceptron network to identify up to ten drivers with an accuracy of 95\% in a controlled field study. 
Additionally they investigate the accuracies of window sizes of 1, 10, 30, and 60 seconds.
Contradicting previous publications, they conclude that a window size of one minute leads to the best result.
In~\cite{Luo2018}, based again on a controlled field study, the authors use a RF algorithm to identify a subset of 4 out of 15 drivers with an accuracy of 89\%.
The results are achieved by tuning the parameters for the number of trees and number of features.

\subsection{Naturalistic Driving Studies}
\label{nat_studies}
The first results using real naturalistic driving data were published in~\cite{Zhang2016} and~\cite{Hallac2016}.
Using a 30-second window for feature extraction, the researchers were able to perfectly distinguish between two drivers~\cite{Zhang2016}.
Accuracies dropped when additional drivers were added to below 40\% for nine and more.
In~\cite{Wang2017}, researchers approached the topic by applying findings and results from~\cite{Enev2016} onto the collected naturalistic driving data with an optimal window length of 5 seconds, which lead to perfect identification of 30 drivers.
However, to achieve this, all available CAN-bus signals and a training set of over four hours and at least ten minutes of driving data for testing was required.
Limitations, according to the authors, of leveraging such a data set were partially missing signals and potential fitting of the model to the cars' differences since each participant was using his/her own car.
``However, the different mileage and ages of vehicles might cause differences in vehicle dynamics.
That means the random forest model might potentially pick up some differences between the vehicles."~\cite{Wang2017}.
Further results on naturalistic driving data is presented in~\cite{Dong2017}.
The authors were able to identify 50 drivers with an accuracy of 40\% by using 160 trips per driver for training and 40 trips for testing.
For that, a special Autoencoder Regularized Network (ARNet) was developed consisting of stacked recurrent neural networks (RNN) using gated recurrent units (GRU) and fully connected (FC) layers with a window size of 4 seconds.
%A window size of 4 seconds was applied on speed norm, difference of speed norm, acceleration norm, difference of acceleration norm, and angular speed, which were derived from the GPS signal.

A fundamentally new approach for driver identification was taken in~\cite{Hallac2016}.
The authors limited themselves only to turning events and tried to identify the drivers leveraging all available CAN-bus signals.
The approach showed promising accuracy improvements over a random guess, however with 50\% accuracy for 5 drivers it did not reach the performance of previous results.
In a later approach, the same research group increased identification rates to 51\% for 56 drivers by applying GRUs with a 1-second window onto the available CAN-bus signals~\cite{Hallac2018}.
Similar to~\cite{Wang2017}, an enormous amount of driving data ($\approx$ 23 hours per driver) was required to train the model.
Following the event-based approach of~\cite{Hallac2016}, the authors of~\cite{Gahr2018a} tried to identify up to 50 drivers by only using the brake pedal signal.
They limited themselves to the brake pedal signal, collected under a naturalistic driving setting, based on the results in~\cite{Enev2016} and showed that the brake pedal indeed serves as a good identifier of the driver.
Additionally they showed that identification performances can be significantly increased to almost perfect identification for 5 and 85\% accuracy for 50 drivers by tailoring a model to one specific signal.
However, they also showed that the brake pedal is only sparsely used over time while driving.

%Conclusion
%Concluding, perfect driver identification was only achieved under controlled data collection settings and using an enormous amount of driving data.
%Further, limiting to specific signals can significantly increase identification accuracies, because models can be tuned to the sensor, such as the brake pedal signal.
%However, the sparse usage of the brake pedal requires a new approach, such as via the steering wheel signal for a constant input from the driver and hence a potential earlier identification.
%Last, the reported window sizes for the models are contradicting in the literature and reach from one second to over a minute.
Concluding, perfect driver identification has only been achieved under controlled data collection settings.
High accuracies were achieved using sensor data from the longitudinal vehicle control (i.e. brake pedal and throttle).
The reported window sizes for the models are contradictory in the literature and range from under one second to over a minute.
Using naturalistic driving data, two key findings were identified from related work.
First, an enormous amount of data is required to achieve sufficiently good results.
And second, if every study participant uses their own car, limiting to a single sensor also limits the chance of misfitting the model onto the car.
Therefore we identified the research gap of driver identification using naturalistic driving data with the limitation on lateral vehicle control (i.e. steering wheel signal).

\section{Data \& Methods}
\label{data}
In the following, we describe our signal processing methods and underlying theoretical approach.
The analysis is based on data collected from a field study under naturalistic driving conditions of 72 patrollers from a road assistance service.
The patrollers were distributed over 10 different base locations in the German-speaking part of Switzerland, covering mountainous as well as hilly to almost flat regions.  
They were not asked to perform any special maneuvers and followed their daily routine.
All but one driver was male, ranging in age from 21 to 64 years old with a mean of 40 years.
Further, every participant was driving a Chevrolet Captiva.
%The study participants received as part of their job a training for safe and eco-friendly driving.
The study ran for approximately five months.
In total, over 680'000 km of driving data was collected by reading out CAN-bus data via the ODB-II port with a dongle that transferred the signals via Bluetooth to a smartphone, where GPS and accelerometer signals were also added and then transmitted in real time via the cellular network to a server.
The 72 drivers transmitted on average over 52 hours of driving data.

\subsection{Preprocessing}
\label{preproc}
Each trip is queried sperately from the database containing the raw collected data.
This means that the recorded data is prone to the inconsistent sampling, Bluetooth data transmission errors from the dongle to the smartphone, and transmission outages from the smartphone via the cellular network to the server.
The trips were queried as a \textit{Pandas} DataFrame\footnote{https://pandas.pydata.org}.
%Due to the potential erroneous trips, the preprocessing is comprised of two parts.
We first cleaned the data by removing all erroneous queried values, such as \textit{NaNs}.
Additionally, all data points with GPS outage were removed.
%After cleansing the queried trip, larger outages were identified. 
%Due to a sampling frequency of approximately 10 Hz of the steering wheel, a trip is treated and hence split into two if the time gap between two recorded values is larger than one second.
%These trips were then forwarded to the data preparation part.

\begin{figure}[h]
\centering
\includegraphics[width=\linewidth]{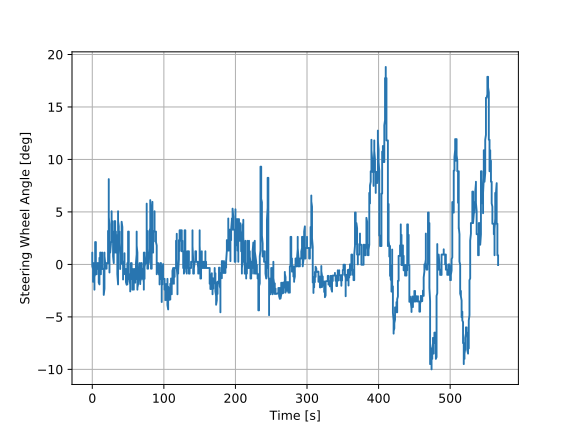}
\caption{Example of a recorded steering wheel signal.}
\label{sw_0}
\end{figure}
In a second step, the data is resampled and quadratically interpolated at a frequency of 10~Hz, consistent with prior research~\cite{Enev2016}.
Last, each of the trips is required to have a minimum length of 5 minutes.
When a preprocessed trip is shorter than the required minimum length, the trip is dismissed.
An example of a recorded and preprocessed signal can be seen in Figure~\ref{sw_0}.
In the following, we describe the signal of each trip by $X_{i,t}$ where $t \in [0,T_i]$ and $T_i$ denotes the total number of samples of the $i$-th recorded trip.
%It is worth noting that two trips most likely have a different length, i.e. $T_i \neq T_j$ for $i \neq j$.
% All trips for the later training, testing and evaluation are stored in a so called ``Signal Container'' (SC).

\subsection{Stationarity \& Correlated Lag}
\label{stationarity}
%In this subsection we highlight our theoretical contribution in this paper.
We motivate the design of our neural net by the analysis performed on the collected steering wheel signals in this section.
In the following we will show that the steering behavior can be modeled as a stationary signal.
This will allow us to draw concrete design decisions for an optimal window length for the models introduced in Section~\ref{model}.

For simplicity, we neglect the trip notation and simply write $X_t$ for an observed signal.
Following the description of~\cite{Fuller1996}, the expected value of the observed signal is denoted by $E\{X\}$ and the covariance of $X_t$ with the signal itself at a different time step $X_{t+h}$ is denoted as $Cov\{X_t, X_{t+h}\} = E\{X_t X_{t+h}\} = \gamma(t,h)$. 
% Short excursion into stationary signals.
In the following we use the definition of~\cite{Fuller1996} of a stationary time series:
\begin{enumerate}
\item the expected value is a constant for all $t$, i.e. $E\{X_t\} = c, \forall t$, and
\item the covariance is time-independent, hence only depends on the lag $h$, or time distance, between two values, i.e $Cov\{X_t, X_{t+h}\} = E\{X_t X_{t+h}\} = \gamma(h),  \forall t$.
\end{enumerate}
Last, we introduce the autocorrelation function
\begin{equation}
\label{eq_acf}
\rho(h) = \frac{\gamma(h)}{\gamma(0)},
\end{equation}
which normalizes the autocovariance function, s.t. the following properties are valid:
$\rho(0)=1$, $|\rho(h)| \leq 1$, and $\rho(h) = \rho(-h)$.

Following~\cite{Fuller1996}, we test for a unit root at a 1\% significance level of all trips\footnote{We used the library http://www.statsmodels.org in it's version 0.9.0 for python3}.
In case the H0 hypothesis of a unit root is rejected, i.e. a p-value below the 1\% significance level is returned, we assume the steering wheel signal to be stationary.
Our analysis showed that over 12'278 tested trips of all 13'883 were returned with a rejected H0 hypothesis.
%Hence we assume that the steering wheel signal can be modeled as a stationary process.
Using the stationary property of a process without a constant frequency, we can further assume a fading ACF, i.e. after some lag $t+h > t+h_{cor}$, the observed measurements are not significantly correlated with the the measured value at $t$.
The property of a fading ACF does also hold for the 1'605 signals where the H0 hypothesis was not rejected, however finding the uncorrelated lag $h_{cor}$ is extremely cumbersome since Equation~\eqref{eq_acf} becomes time-dependent and hence the lag has to be found at each sampling step.

\begin{figure}[h]
\centering
\includegraphics[width=\linewidth]{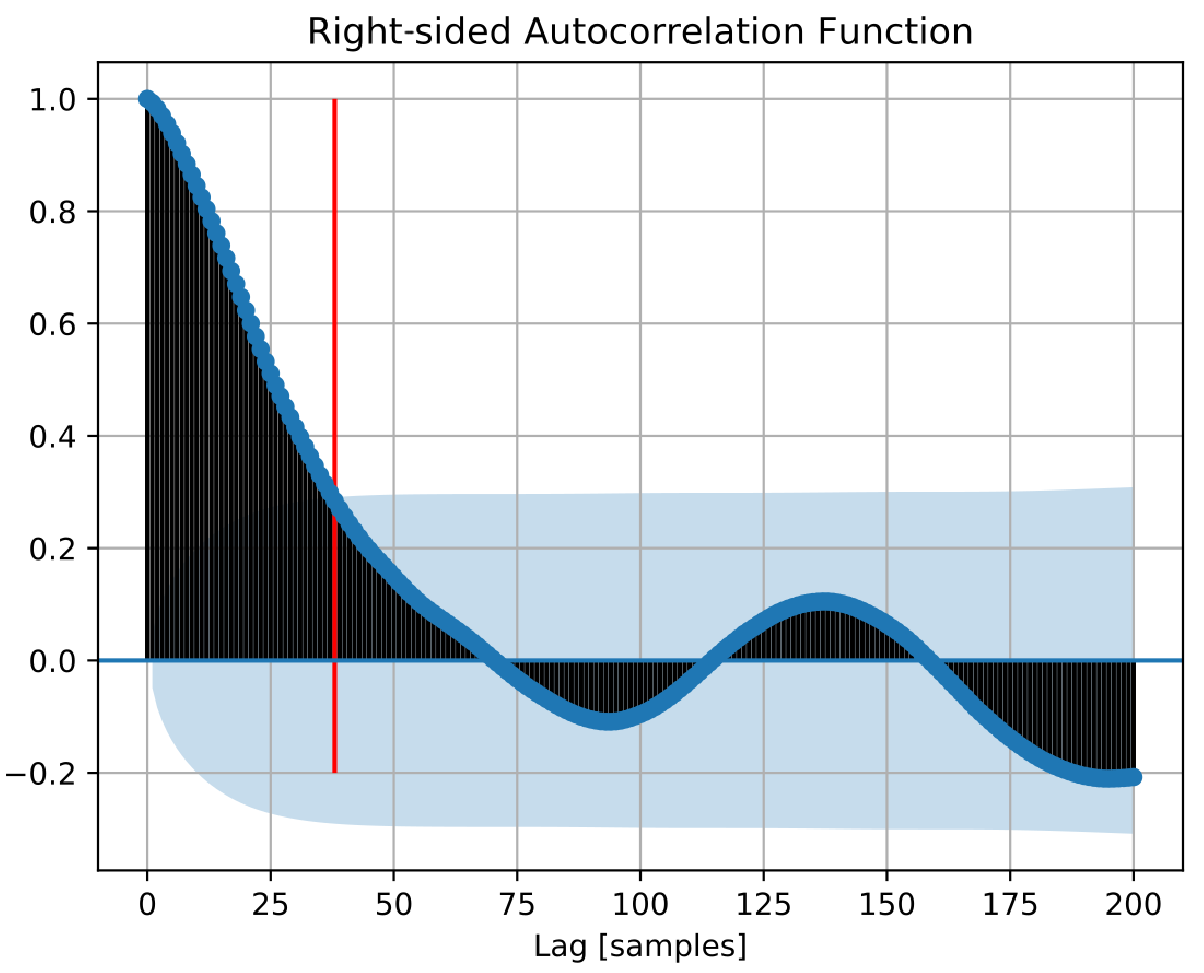}
\caption{Example of a right-sided autocorrelation function and the 1\% confidence region.}
\label{acf_0}
\end{figure}

Due to the symmetry property, it is sufficient to only look at the right-sided ACF.
An example of the right-sided ACF of a recorded steering wheel signal is shown in Figure~\ref{acf_0}.
In the plot we can see that the correlation of a signal value at time lag $h$ fades out and stays within a significantly uncorrelated region, as the lag $h$ gets larger.
The lag at which the ACF and the significance region intersect ($h_{cor}$) is displayed in Figure~\ref{acf_0} by the vertical red line, while the significance region is shown in light blue.

\begin{figure}[h]
\centering
\includegraphics[width=\linewidth]{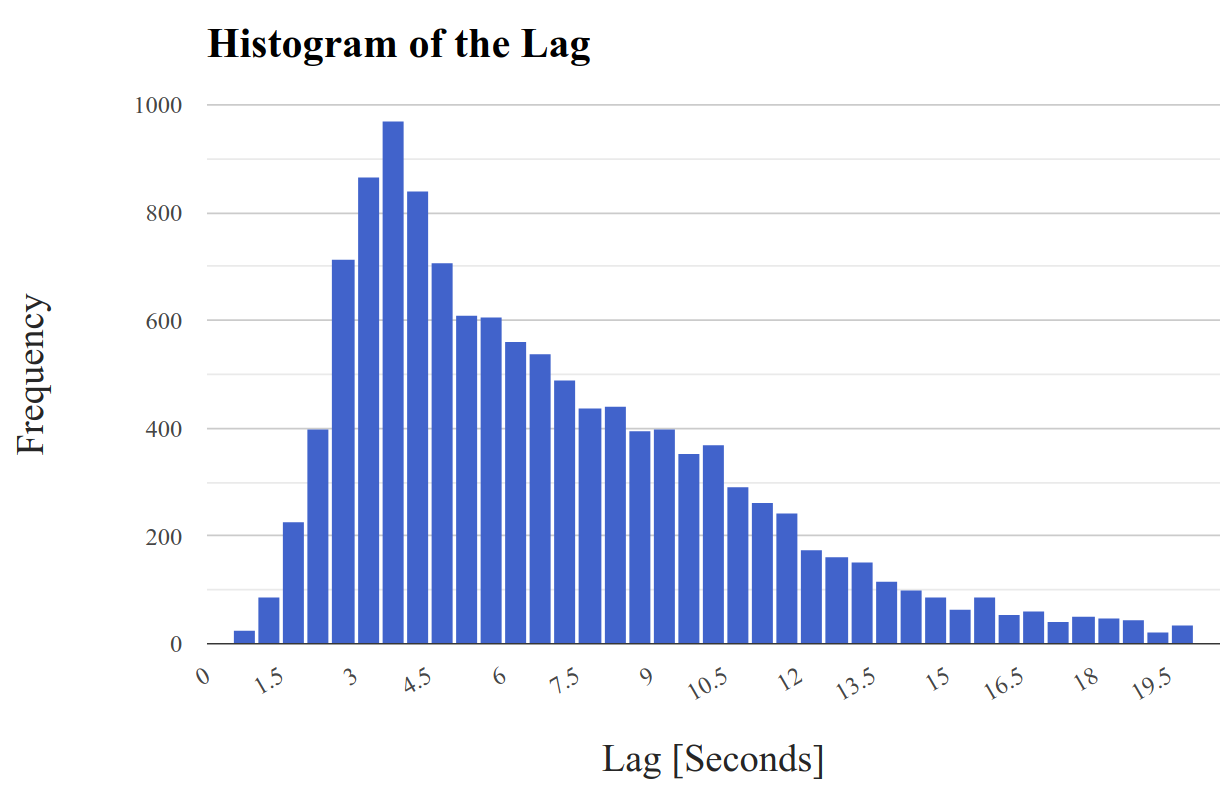}
\caption{Distributions of the maximum lags for all stationary trips.}
\label{lag_dist}
\end{figure}

For the design of our neural net, we want to select a window size that corresponds to a $\hat{h}_{opt}$ which is large enough so that all significantly correlated values with lag $h_{cor,i}$ are still taken into account.
At the same time, we want to avoid $\hat{h}_{opt}$ being too large, so that uncorrelated (noise-contributing) values are taken into account.
Following this approach, we analyzed all stationary trips.
Doing that, we limited the maximum lag to 20 seconds.
For the analysis we required again a significance level of 1\%.
Figure~\ref{lag_dist} shows the histogram of all found lags.
The mode of the histogram lies at 3.6 seconds, while the median and mean values lie at 5.8 seconds and 6.6 seconds, respectively.
This indicates that an optimal window length should also lie in the range of 2 to 10 seconds.
In Chapter~\ref{res_acf} we validate these findings and present an optimal window size $\hat{h}_{opt}$.

%The distribution shown in Figure~\ref{lag_dist} indicate that the optimal window length should lie between the minimal value of all found lags $h_{min} =\displaystyle \min_{i}(h_i)$ at 0.5 seconds and the maximum at 20 seconds.
%However, since the histogram has a positive skew, we expect an optimal window size towards the lower range around the mode at 3.6 seconds, to the median at 5.8 seconds or the mean at 6.6 seconds.
%For larger $\hat{h}_{opt}$ the noise contribution of trips with a smaller $h_i$ will be too strong and accuracies are expected to drop.

\subsection{Method Description}
\label{method}

In the following we describe the signal's postprocessing and feeding into the neural network.
In contrast to previous attempts using GRUs~\cite{Hallac2018}, we do not rely on the raw signals but compute the logarithmic spectrogram as additional features and feed them into the network to improve the accuracy~\cite{Zihlmann2017}.
 
\subsubsection{Feature Preparation}
\label{features}
We first extract segments of a predefined length from each trip.
Since the trips  might vary in length, this operation helps ensure a fixed length.
In our case, we define the length of a segment to be $S = 15$~min.
Multiple segments can be extracted from one trip, if $T_i>S$, and one segment can contain multiple trips, if $S>T_i$, by simply concatenating the trips.
We make sure every driver has the same number of segments.
This way, we ensure a balanced dataset for training and testing.
For each segment, we apply a sliding window $\vec{W}_w$ without overlapping to extract features.
The length of the sliding window relates the previously found $\hat{h}_{opt}$.
We denote these windows as $\vec{W}_w[t]$ with $t \in [0, \hat{h}_{opt}]$.
%It is worth to note that $\hat{h}_{opt} \ll S$. 
Each of the windows is then used to calculate one feature vector $\vec{F}_w$.
We chose a rather simple feature representation of the signal and only took the logarithmic Fourier-Transformation (LFT) of a window:
\begin{equation}
\label{eq:feature_eng}
\vec{F}_w = \log_2(|\mathrm{FFT}(\vec{W}_w)| + 1),
\end{equation} 
where $\mathrm{FFT}(\cdot)$ denotes the Fast-Fourier-Transformation and hence, $\vec{F}_w$ is of length $\hat{h}_{opt}$.

Additionally to the LFT, the average velocity of the car in each window was appended to $\vec{F}_w$.
The motivation for the velocity is twofold.
First, the transmission from the steering wheel angle to the actual wheel angle --- and hence the curve radius --- is nonlinear in modern cars.
To avoid dangerous strong steering at high speeds, which could lead to a potential rollover of the car, the steering wheel has a lower angle transmission towards the wheels at those speeds~\cite{Koehn2004}.
Second, the centripetal force is linearly dependent on the velocity.
As the lateral force is a major feedback for the driver on the driving style, we assume that it is also a good identifier.
However, to avoid identification of the driver via longitudinal driving behavior, we restricted ourselves to only the mean velocity of one window.
The feature calculation hence results in a vector $\vec{F}_w = [f_{w}[0], \hdots, f_{w}[f], \hdots, f_{w}[\hat{h}_{opt}]]^{T}$ of length $\hat{h}_{opt}+1$ for each window $\vec{W}_w$.
For each segment, a sequence of feature vectors is obtained and results in one segment matrix per class (i.e. driver) $i$:
\begin{align}
\label{eq:seg_feature}
\mat{B}_i &= 
\begin{bmatrix}
\vec{F}_1 
\hdots 
\vec{F}_w
\hdots 
\vec{F}_F
\end{bmatrix}\nonumber\\
&=
\begin{bmatrix}
&f_{1}[0] &\hdots &f_{w}[0] &\hdots &f_{F}[0]\\
&\vdots  &&\vdots  &&\vdots\\
&f_{1}[f] &\hdots &f_{w}[f] &\hdots &f_{F}[f]\\
&\vdots  &&\vdots  &&\vdots\\
&f_{1}[\hat{h}_{opt}] &\hdots &f_{w}[\hat{h}_{opt}] &\hdots &f_{F}[\hat{h}_{opt}]
\end{bmatrix},
\end{align}
where $F$ is the number of windows within one segment, i.e. $F = S/\hat{h}_{opt}$.
The segment matrices were concatenated so that the model was fed by a [32, $F$, $\hat{h}_{opt}$] tensor.
We chose a batch size of 32~\cite{Masters2018}, i.e. 32 segment matrices of uniformly randomly picked drivers were used to create one full batch matrix.

\subsubsection{Model}
\label{model}
The neural network was built using the \textit{tensorflow} library\footnote{We used the python3 library of tensorflow in it's versions 1.13.1, https://www.tensorflow.org/}.
The network processed the features of the windows sequentially and put them into a two-layer bidirectional recurrent neural network (RNN) initialized with GRU cells of size 512.
A GRU cell consists of an update gate ($z_j$) and a reset gate ($r_j$) that update given input $\vec{F}_w$ and hidden state $h_i$ as
\begin{align}
z_j &= \sigma\left([\mat{W}_z \vec{F}_w]_j + [\mat{U}_z \vec{h}_{i-1}]_j \right),\\
r_j &= \sigma\left([\mat{W}_r \vec{F}_w]_j + [\mat{U}_r \vec{h}_{i-1}]_j \right),
\end{align}
Where $\mat{W}$ and $\mat{U}$ are weight matrices which are learned and $\sigma$ is the logistic sigmoid function~\cite{Cho2014}.
The hidden state is then updated by:
\begin{align}
h_{i,j} =& z_{j} h_{i-1,j} + (1 - z_{j}) \cdot\nonumber\\
	&\sigma\left( [\mat{W} \vec{F}_w]_j + \left[\mat{U} (\vec{r} \odot \vec{h}_{i-1})\right]_j\right),
\end{align}
where $\odot$ denotes the Hadamard-, or element-wise product.

\begin{figure}[h]
	\centering
	\includegraphics[width=\linewidth]{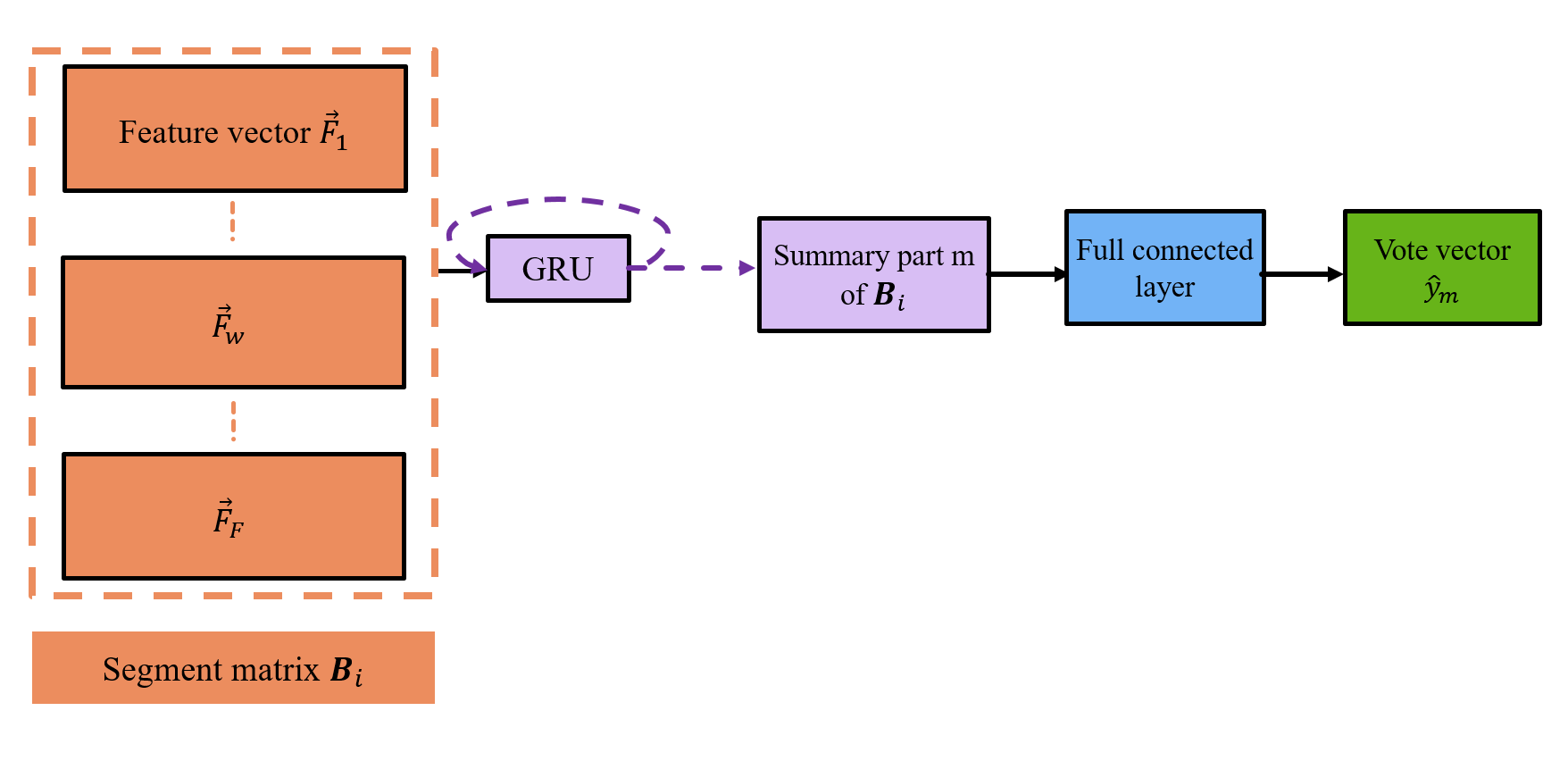}
	\caption{Scheme of our neural network invoking voting mechanism}
	\label{network}
\end{figure}

The 512 outputs of the first GRU are fed into the second GRU stage, returning also 512 outputs.
It should be emphasized that we do not further process the outputs of GRU.
Instead, we use GRU as an encoding method and treat the hidden states of both GRU directions as a summary of a segment, namely the summary of a driver's steering wheel behavior.
These hidden states are then reduced by a fully connected (FC) layer to a softmax vector of length of the number of classes.
We name the softmax vector the \textit{vote vector}.
In addition, to boost the performance of the network, we invoke a voting mechanism in our model, which is inspired by~\cite{Enev2016} and~\cite{Gahr2018a}.
Conventionally, in classification tasks a RNN or its variants (GRU, LSTM) only generates one decision at the end of a input sequence, which could be vulnerable due to vanishing and exploding gradient problems and hence lead to a decision that biases towards the most recent inputs.
Gradient vanishing and exploding problem can be partially solved by GRU- or LSTM variants of RNNs~\cite{Goodfellow2016}.
The scheme of our neural network is provided in Figure~\ref{network}.

As mentioned in the previous paragraph, the hidden states of the GRU, which are reduced to vote vectors, are further processed by a voting mechanism.
In the testing and training phase, for every 6 windows, a vote vector is generated from current hidden states of the network.
%This can be understood as a Bag-of-Words representation.
We use only 6 windows to avoid overfitting to the traffic or road topology~\cite{Dong2017}.
These vote vectors are compared with the ground-truth one hot vector $\vec{y_i}$ with cross entropy as the loss function.
Hence, $F$ feature vectors %$\vec{F}_w$
 in one segment matrix (c.f. Equation~\eqref{eq:seg_feature}) lead to $M = \frac{F}{6}$ vote vectors.
The loss of the vote vectors is then aggregated and the parameters of the network are updated by back-propagation.
The training was done using the RMSPropOptimizer with a learning rate of $10^{-4}$, a state keep probability for the GRUs of 0.7, and a L2-norm weight regularization to avoid overfitting of $\lambda=10^{-3}$.

%((((For evaluation the generated one hot vector is compared through the cross entropy with the actual  one hot vector $\vec{y_i}$.
%\begin{figure}[h]
%\centering
%\includegraphics[width=\linewidth]{GRU_plot}
%\caption{Scheme of our neural network invoking voting mechanism}
%\label{network}
%\end{figure}
%A schematic of the network can be seen in Figure~\ref{network}.))))

\section{Results}
\label{res}
In this section we present the results that we achieved using the neural network from Section~\ref{model}.
First we will give results motivating our approach to design the window size of a neural network via the outfading ACF and the distribution of the lags $h_{cor}$.
Afterwards, the dependency of the model's accuracies and the required observation time are presented.
Finally we investigate the confusion matrix of the model's identification.

We required at least 240 minutes of driving data for each driver to train the model and at least 30 minutes of driving data for testing.
The testing and training data did not overlap.
Through the segment matrix representation, we ensure the same amount of training and testing data for each class.

\subsection{$\hat{h}_{opt}$ - Optimal window size}
\label{res_acf}
\begin{figure}[h]
\centering
\includegraphics[width=\linewidth]{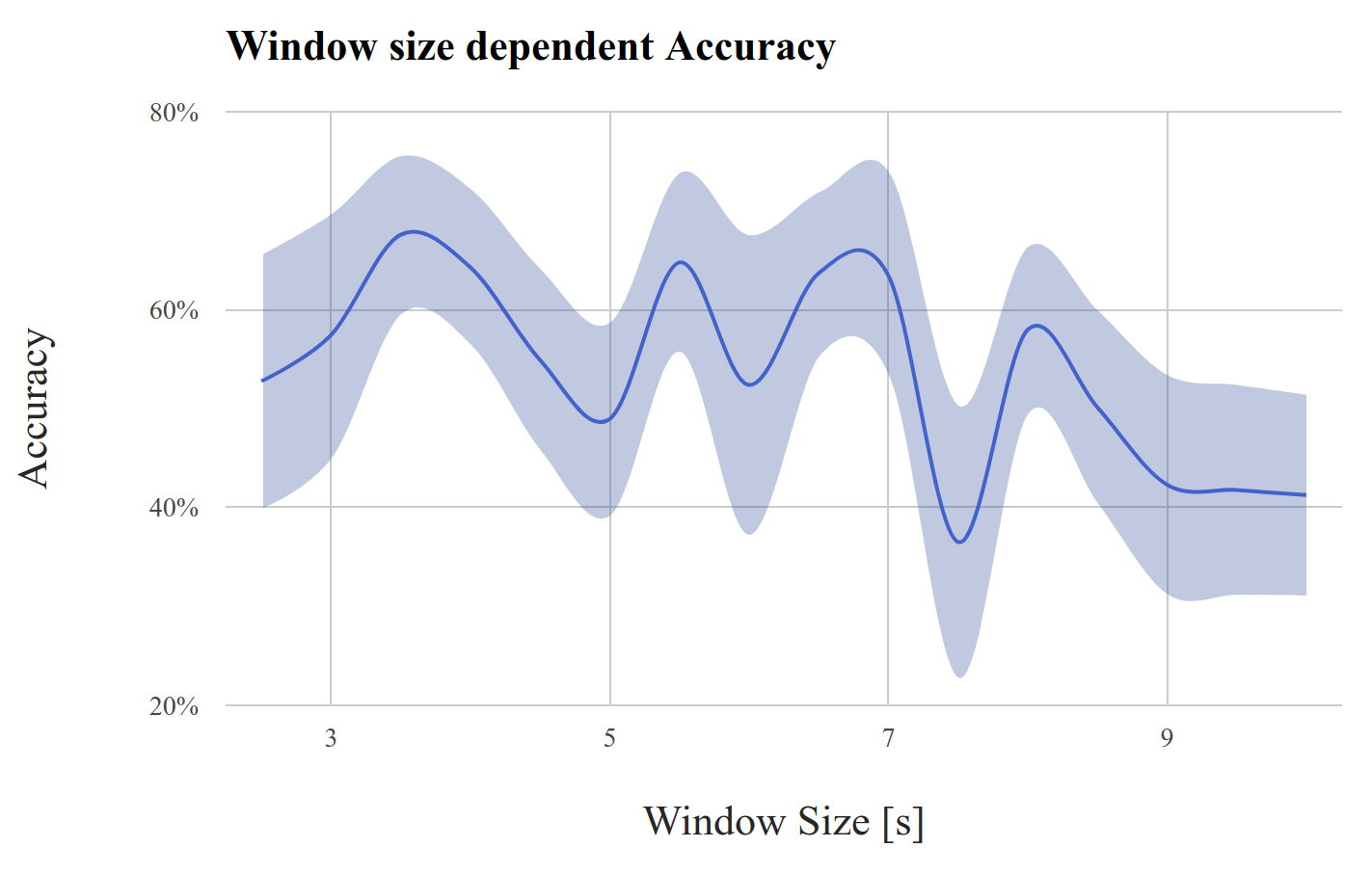}
\caption{Accuracy of our neural net for different window sizes for 15 drivers.}
\label{window_comp}
\end{figure}

We limited the number of drivers to the most common class size of 15 and a testing time of 15 minutes.
For each window size, at least seven sets of drivers were picked at random from our dataset.
Additionally, we used the same set of randomly chosen drivers and trips for each setting to ensure reproducibility.
Since the lag $h$ is distributed from 0.5 seconds to 20 seconds, with a mean of $\approx 5.8$ seconds, we analyzed the performance of our network with window sizes between 2.5 and 10 seconds.
In Figure~\ref{window_comp}, the accuracies and standard deviations are shown.
Additionally to the average accuracy for each window size, the standard deviation is given by the light blue region.
It can be seen that the accuracy has its peak around the mode of the histogram of the lag $h$ at 3.5 to 4 seconds with approximately 67\%.
Although high accuracies of over 60\%, on average, were also achieved for 5.5, 6.5 and 7.0 seconds, the best performance was at 3.5 seconds.
Above 7 seconds, the accuracies drop significantly to below 60\%, hence we limited our analyses to a maximum window size of 10 seconds.
Similar significant drops of accuracies were observed for window sizes below 3.5 seconds.
Window sizes below 2.5 seconds were not considered as they lead to an increased size of segment matrices and calculations become too complex.

\subsection{Replication of the state of the art}
\begin{figure}[h]
\centering
\includegraphics[width=\linewidth]{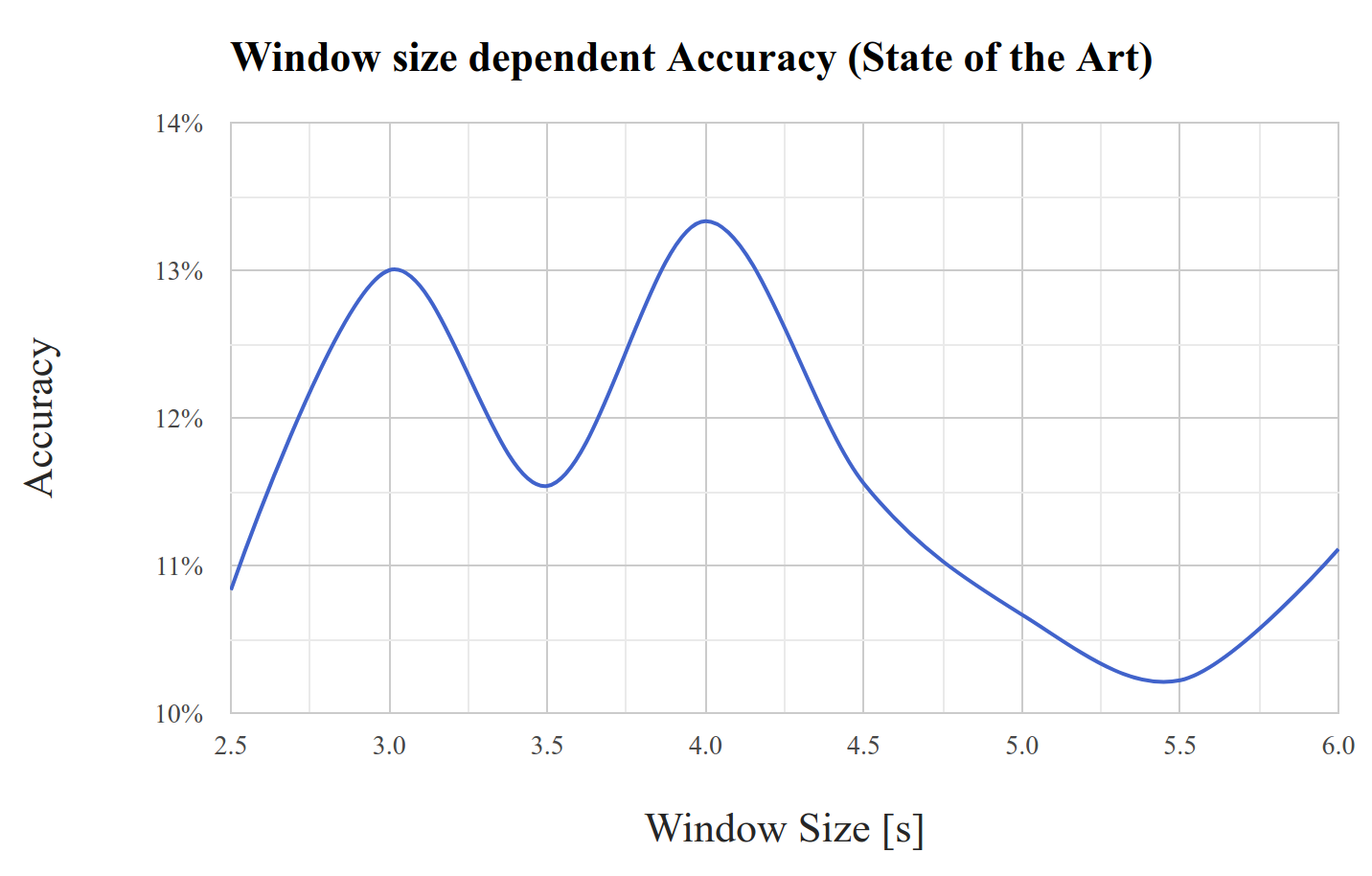}
\caption{Accuracy of the replicated state of the art models for different window sizes for 15 drivers.}
\label{window_comp_enev}
\end{figure}

Besides evaluating the newly introduced neural net, we further replicated the work presented in~\cite{Enev2016}.
The remaining major differences are a lower resampling frequency of 10 Hz due to a lower sampling rate at the CAN-bus and our naturalistic driving data set compared to the controlled driving data set.
The strong claims of 83\% accuracy after a training phase of 90 minutes of driving data per user~\cite{Enev2016} could not be replicated in our setting.
Figure~\ref{window_comp_enev} shows the accuracies of the replicated studies using 240 minutes of driving data per class.
It can be seen that the baseline of a random guess at 7\% was improved, and the best performance was achieved with a window size of approximately 4 seconds similar to results presented in Figure~\ref{window_comp}.
However, with an accuracy of less than 15\%, the difference to our introduced model is significantly lower.
This is in line with previous replicated research.
In~\cite{Gahr2018a}, a difference of over 55\% between naturalistic- and controlled driving data is presented.
A similar difference of approximately 50\% can be seen comparing Figure~\ref{window_comp} and~\ref{window_comp_enev}.

\subsection{Model performance dependency on observation time}
\label{acc}
\begin{figure}[h]
\centering
\includegraphics[width=\linewidth]{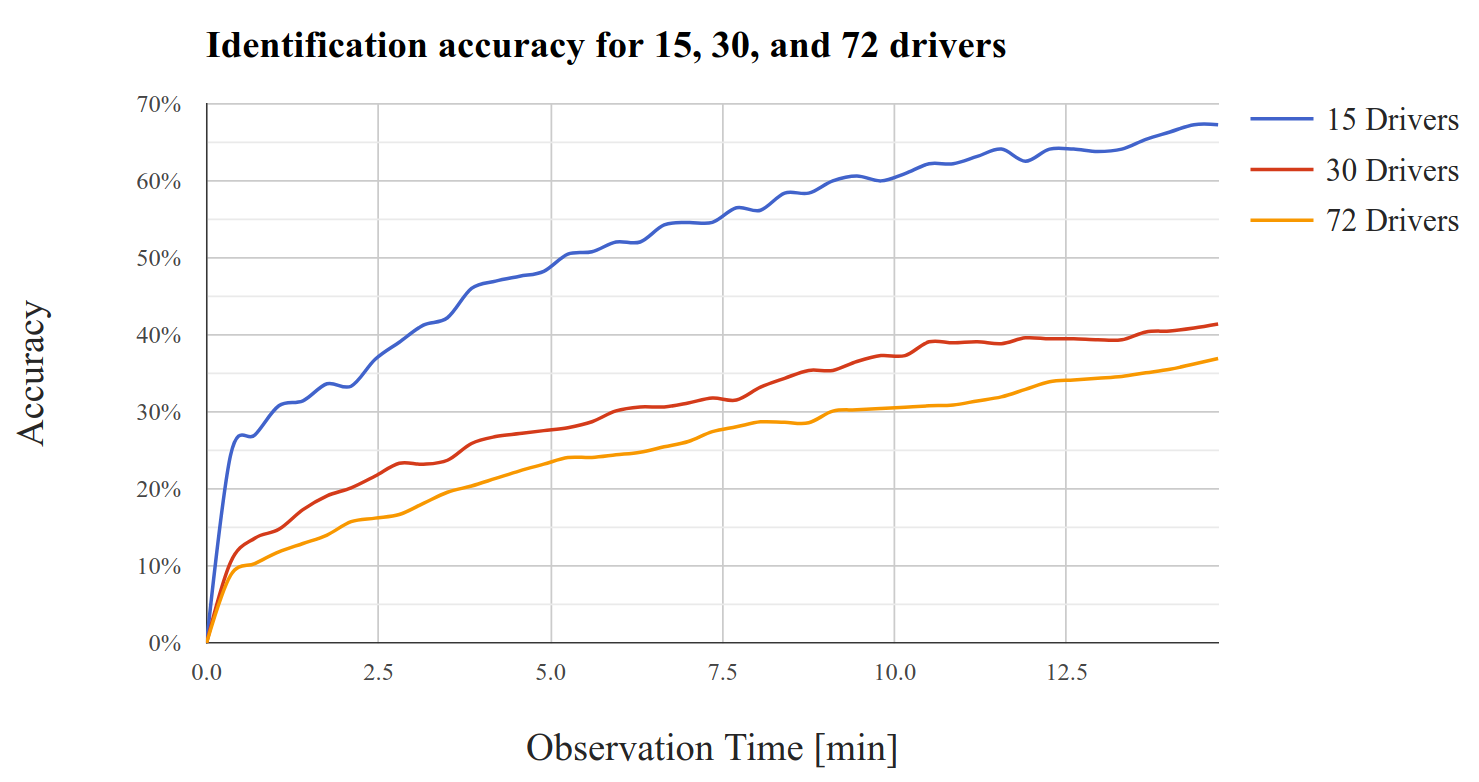}
\caption{Accuracy of our neural net for 15, 30, and 72 drivers over observation time.}
\label{acc_fig}
\end{figure}

In the following we use the optimal window size of 3.5 seconds found in the previous results.
Further we used a training set size of approximately 240 minutes per driver following~\cite{Wang2017}.
We investigate the dependency of the identification accuracy and the observed driving time.
The observation time is limited to 15 minutes, since a longer observation is impractical.
In Figure~\ref{acc_fig} the identification accuracy is shown for a different number of drivers ranging from 15 to the full dataset of 72 drivers.
As expected we see that the accuracy increases with a smaller set size of drivers.
Further, a clear dependency of observation time and accuracy can be observed.
The longer the neural network ``observes'' a driver, the higher the identification accuracy.
Further we can observe that the accuracy tends to stabilize after 12 minutes for 15 and 30 drivers, but not for 72 drivers, indicating that a larger evaluation set size is required for a larger class size to reach a final vote.
After the first vote vector (21 seconds), accuracies lie at approximately 25\% for 15 drivers, 10\% for 30\% for 30 drivers, and 8\% for 72 drivers.
These accuracies increase to 76\% for 15 drivers, increasing the state of the art by a factor of 5 and random guess by a factor of 11 after 42 vote vectors, or almost 15 minutes.
For larger class sizes the accuracies drop to 41\% (30 drivers) and 36\% (72 drivers).

%\subsection{Accuracy dependency on longitudinal vehicle control}
%\begin{figure}[h]
%\centering
%\includegraphics[width=\linewidth]{accuracy_15_4w}
%\caption{Accuracy of our neural net for 15 drivers using all wheel speeds.}
%\label{acc_4w}
%\end{figure}

\subsection{Confusion Matrix}
\label{conf_mat}
\begin{figure}[h]
\centering
\includegraphics[width=\linewidth]{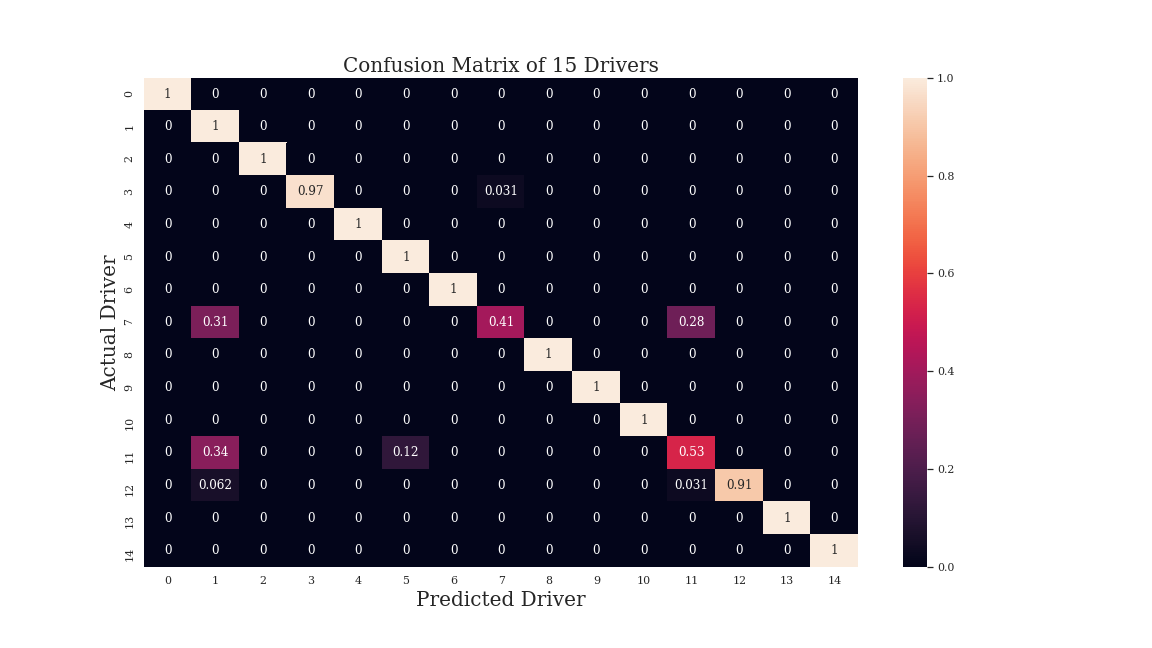}
\caption{Confusion matrix of our neural network for 15 drivers.}
\label{conf_mat_fig}
\end{figure}

Figure~\ref{conf_mat_fig} shows the normalized confusion matrix of 15 drivers after training the algorithm with 240 minutes of driving data averaged over 315 different 15-minute segments. (i.e. 21 segments per driver)
We see that three of the 15 drivers were identified correctly for each of the 21 segments.
It can also be seen that for 12 of the 15 drivers, the diagonal element is the strongest element, indicating a major correct identification.
The remaining three drivers (Driver 5, 8, and 12) hence are on average mostly misidentified.
As described in Section~\ref{data}, the drivers were stationed in different regions with a strongly varying road topology.
Only Driver 12 was extremely misidentified towards Driver 9.
However, the two drivers were stationed at different locations, therefore similar steering behavior is more likely than misfitting of the model towards the region.
Finally we can observe that the neural network was strongly misfitted towards Driver 10, creating a ``sink.''
Again, none of the misidentified drivers were stationed in the same region as Driver 10, therefore a similar driving style is most likely the reason for this over-fitting.
We neglected the confusion matrix for 72 drivers in this section for readability reasons.
The same effects (with lower accuracies, c.f. Figure~\ref{acc_fig}) can be observed.

\section{Discussion and Outlook}
%Intro
Research into the growing field of automated vehicle person identification is motivated by various use-cases, including privacy, personalized service models, and the potential to enable advanced applications such as health and well-being related products and services.
The results presented in this work contribute to the understanding of the current status of driver identification systems.

% Limitations
The results should be seen in the light of the study limitations.
Despite the unique size of the dataset, we acknowledge that the results are based on a naturalistic field test of professional road assistance drivers.
As such, generalization of the results remains a core challenge that should be addressed in future work.
However, we expect that there is reduced variance in the feature distribution of professional drivers when compared to a sample of more ordinary drivers.
As the homogeneity of the sample may have limited the prediction results, greater identification accuracy might be possible in a more general setting.
Thus, we encourage other researchers to validate the strong claims made in this work on more generalizable data sets.
Further, as we limited ourselves to an extremely low dimension of features, only using the logarithmic Fourier-transformation of the steering wheel and the average velocity-per-window, more precise and faster identification could be expected with greater feature engineering.

% Contributions
In summary, this work is based on an extensive literature review that, to our knowledge, has not been previously conducted by the field.
We identified driver identification based on naturalistic driving data using only steering wheel signal as a research gap.
Large discrepancies in literature on the choice of the window size motivated our signal-processing-based approach of an optimal hyper parameter choice.
The newly introduced approach is a lightweight method, which could find further applications outside this field in time-series-dependent tasks.
Additionally, we present a GRU-based neural network identification setup that improves the results from state-of-the-art methods by a factor of almost 5.
The analysis goes beyond the traditional sample size of previous studies, and considers the largest fleet-based setting of 72 drivers under naturalistic conditions.
Finally, to consolidate our contribution, we investigated the confusion matrix and confirm that the model did not have bias toward the region of driving.
Rather, drivers were misidentified because of their similar driving style, suggesting that a more heterogeneous driver set would lead to higher accuracies.

%Outlook
Based on the improvements achieved, we conclude that single-sensor-tailored approaches can increase accuracies significantly, while at the same time avoid misidentification bias from the car or the region.
Hence, in future work, this approach should be combined with existing methods, and for additional sensors, in order to leverage the maximum results and demonstrate the optimal real-world capabilities of data enabled driver identification.

\appendices
%\newpage
\begin{landscape}

\begin{table}
\fontsize{7}{11}\selectfont
\centering
\begin{tabular}{| l | c | c | c | c | c | c | c |} 
\hline
Author & Test Environment &
\makecell{ML Method} &
Window Size & 
Algorithm Input & 
Training Set &
Class Size &
Accuracy\\
\hline\hline

\multirowcell{3}{Gahr et al. 2018~\cite{Gahr2018a}} &
\multirowcell{3}{\shortstack{Naturalistic driving}} &
\multirowcell{3}{\shortstack{Random-forest}} &
\multirowcell{3}{\shortstack{Event based}} &
\multirowcell{3}{\shortstack{CAN-bus Brake pedal signal}} &
%\multirowcell{3}{\shortstack{\\min, max, mean, quartiles, \\
%std dev, autocorr., kurtosis, skew,\\
% 10-PAA, Frequency Components,\\
% via Brake Events
%}} &
\multirowcell{3}{\shortstack{120 - 800 brakings \\$\approx$ 50min - 5h30min\\per driver}} &
 5 & 95.0\% - 99.5\% \\ \cline{7-8}
&&&&&& 15 & 80.0\% - 92.0\% \\ \cline{7-8}
&&&&&& 50 & 70.0\% - 85.0\% \\ 
\hline

\multirowcell{1}{Hallac et al. 2018~\cite{Hallac2018}} &
\multirowcell{1}{Naturalistic driving} &
\multirowcell{1}{\shortstack{Gated recurrent units}} &
\multirowcell{1}{\shortstack{1 Second}} &
\multirowcell{1}{\shortstack{ 665 CAN-bus signals}} &
%\multirowcell{1}{\shortstack{ \\\\Normalized and\\ resampled signals}} &
\multirowcell{1}{\shortstack{$\approx$ 23 hours per driver}} &
\multirowcell{1}{ 56 } &
\multirowcell{1}{ 51.3\% }\\
\hline

\multirowcell{4}{Hallac et al. 2016~\cite{Hallac2016}} &
\multirowcell{4}{Naturalistic driving} &
\multirowcell{4}{Random-forest} &
\multirowcell{4}{\shortstack{Event based}} &
\multirowcell{4}{\shortstack{12 CAN-bus \\ signals}} &
%\multirowcell{4}{\shortstack{ min, max, mean,\\
%	quartiles, std dev,\\
%	autocorr., kurtosis, skew\\
%	5-PCA of DWT\\
%	via Turning Events
%	}} &
\multirowcell{4}{ \shortstack{11-34 turnings \\ per driver}} &
 2 & 76.9\% \\ \cline{7-8}
&&&&&& 3 & 59.4\% \\ \cline{7-8}
&&&&&& 4 & 55.2\% \\ \cline{7-8}
&&&&&& 5 & 50.1\% \\ 
\hline

\multirowcell{2}{Wang et al. 2017~\cite{Wang2017}} &
\multirowcell{2}{\shortstack{Naturalistic driving}} &
\multirowcell{2}{\shortstack{Random-forest}} &
\multirowcell{2}{\shortstack{5 Seconds}} &
\multirowcell{2}{\shortstack{CAN-bus Signals: Speed,Torque, \\
				Brake Pressure,	Throttel Position, Steering Angle}}&
%				Derived Signals: Long. Acc.,\\
%				Long. Jerk, Steering Speed}} &
%\multirowcell{1}{\shortstack{\\Mean, Minimum, \\
%							Maximum, 85\textsuperscript{th} Percentile, \\
%							Standard Deviation,\\
%							via forward selection method\\
%							and sliding window}} &
\multirowcell{2}{\shortstack{4 hours per driver}} &
\multirowcell{2}{30 } &
\multirowcell{2}{93\% - 100\% }\\[2ex]
\hline

\multirowcell{1}{Dong et al. 2017~\cite{Dong2017}} &
\multirowcell{1}{\shortstack{Naturalistic driving}} &
\multirowcell{1}{\shortstack{ARNet}} &
\multirowcell{1}{\shortstack{4 Seconds}} &
\multirowcell{1}{\shortstack{6 GPS-speed derived Signals}}& %: Speed,\\
%				speed norm, difference of speed norm,\\
%				acceleration norm, difference of acceleration norm, and angular speed}} &
%\multirowcell{1}{\shortstack{mean, min, max,\\
%				25\%, 50\% and 75\% quartiles,\\
%				standard deviation}} &
\multirowcell{1}{160 trips} &
\multirowcell{1}{50 } &
\multirowcell{1}{40\% }\\
\hline

\multirowcell{1}{\shortstack{Zhang et al. 2016~\cite{Zhang2016}}} &
\multirowcell{1}{\shortstack{Naturalistic driving}} &
\multirowcell{1}{\shortstack{SVM}} &
\multirowcell{1}{\shortstack{30 Seconds}} &
\multirowcell{1}{\shortstack{CAN-bus \& Smartphone Data}} &
%\multirowcell{1}{\shortstack{\\a}} &
\multirowcell{1}{\shortstack{$\approx$1.25 hours}} &
\multirowcell{1}{\shortstack{14}} &
\multirowcell{1}{\shortstack{29\%}}\\
\hline

\multirowcell{2}{Wakita et al. 2005~\cite{Wakita2005}} &
\multirowcell{1}{\shortstack{Simulator study}} &
\multirowcell{2}{\shortstack{\\Gaussian Mixed \\ Model (GMM) }} &
\multirowcell{2}{\shortstack{0.6 Seconds}} &
\multirowcell{2}{\shortstack{\\CAN-bus signals \\ \& following distance}} &
%\multirowcell{2}{\shortstack{\\unclear, \\ not well described}} &
\multirowcell{2}{10 min} &
\multirowcell{1}{ 12 } &
\multirowcell{1}{ 81\% }\\ \cline{2-2}\cline{7-8}
&\multirowcell{1}{\shortstack{\\Controlled driving study}}&&&&&
\multirowcell{1}{30} & \multirowcell{1}{73\%} \\
\hline

\multirowcell{2}{Miyajima et al. 2006~\cite{Miyajima2006, Miyajima2007}} &
\multirowcell{1}{\shortstack{Simulator study}} &
\multirowcell{2}{\shortstack{\\GMM}} &
\multirowcell{2}{\shortstack{0.8 Seconds}} &
\multirowcell{2}{\shortstack{\\CAN-bus signals \\ \& following distance}} &
%\multirowcell{2}{\shortstack{unclear,\\ via windowing}} &
\multirowcell{2}{3 min} &
\multirowcell{2}{276} &
\multirowcell{1}{ 89.6\% }\\ \cline{2-2}\cline{8-8}
&\multirowcell{1}{\shortstack{\\Controlled driving study}}&&&&&
 & \multirowcell{1}{76.8\%} \\
\hline

\multirowcell{1}{\shortstack{Del Campo et al. 2014~\cite{Martinez2014}}} &
\multirowcell{1}{\shortstack{Controlled driving study~\cite{UYANIK2008}}} &
\multirowcell{1}{\shortstack{Multi-layer Perceptron}} &
\multirowcell{1}{\shortstack{2 Seconds}} &
\multirowcell{1}{\shortstack{Gas- \& Brake pedal CAN-bus signal}} &
%\multirowcell{1}{\shortstack{Cepstral Analysis}} &
\multirowcell{1}{\shortstack{$\approx$ 8 min}} &
\multirowcell{1}{\shortstack{3 - 5 }} &
\multirowcell{1}{\shortstack{84\% - 70\%}}\\
\hline

\multirowcell{1}{\shortstack{Martinez et al. 2016~\cite{Martinez2016}}} &
\multirowcell{1}{\shortstack{Controlled driving study~\cite{UYANIK2008}}} &
\multirowcell{1}{\shortstack{Extreme Learning Machine}} &
\multirowcell{1}{\shortstack{128 Seconds}} &
\multirowcell{1}{\shortstack{14 CAN-bus signals}} &
%\multirowcell{1}{\shortstack{Cepstral analysis}} &
\multirowcell{1}{\shortstack{N.A.}} &
\multirowcell{1}{\shortstack{2 - 11}} &
\multirowcell{1}{\shortstack{97\% - 85\%}}\\
\hline

\multirowcell{6}{Enev et al. 2016~\cite{Enev2016}} &
\multirowcell{6}{\shortstack{Controlled driving study}} &
\multirowcell{6}{Random-forest} &
\multirowcell{6}{3.5 Seconds} &
	\multirowcell{2}{\shortstack{15+ CAN-bus \\ signals}} &
%\multirowcell{6}{
%\shortstack{\\min, max, mean,\\
%quartiles, std dev,\\
%autocorr., kurtosis,\\
%skew, 10-PAA, \\
%Frequency Components,\\
%via Sliding Windows
%}} &
90 min & 
\multirowcell{6}{15} &
		100\% \\ \cline{6-6}\cline{8-8}
&&&&&	15 min & & 100\% \\ \cline{5-6}\cline{8-8}
&&&&	\multirowcell{2}{Brake Pedal Signal} 
&		90 min & & 100\% \\ \cline{6-6}\cline{8-8}
&&&&&	15 min & & 87.33\% \\ \cline{5-6}\cline{8-8}
&&&&	\multirowcell{2}{Steering Wheel Signal} 
&		90 min & & 83.33\% \\\cline{6-6}\cline{8-8}
&&&&&	15 min & & 64.67\% \\ 
\hline

\multirowcell{1}{\shortstack{Burton et al. 2016~\cite{Burton2016}}} &
\multirowcell{1}{\shortstack{Simulator study}} &
\multirowcell{1}{\shortstack{Decision Trees, SVM, kNN}} &
\multirowcell{1}{\shortstack{10 Seconds}} &
\multirowcell{1}{\shortstack{Vehicle coordinates \& CAN-bus signals}} & %speed,steering \\
%				  wheel, throttle, brake pedal}} &
%\multirowcell{1}{\shortstack{\\Mean and std. dev.\\derivations of the signal}} &
\multirowcell{1}{\shortstack{16 min}} &
\multirowcell{1}{\shortstack{10}} &
\multirowcell{1}{\shortstack{100\%}}\\
\hline

\multirowcell{3}{Jafarnejad et al. 2017~\cite{Jafarnejad2017}} &
\multirowcell{3}{\shortstack{\\Controlled driving study}} &
\multirowcell{3}{\shortstack{Random-forest \\
				w/ variations \\
				\&	SVM}} &
				
\multirowcell{3}{\shortstack{15 Seconds}} &
\multirowcell{3}{\shortstack{CAN-bus Signals:
				Speed, Torque,\\ Throttel,
				Yaw Rate, Steering Angle}} &
%\multirowcell{3}{\shortstack{\\minimum,
%				maximum, mean,\\
%				median, standard deviation,\\
%				kurtosis, skewness\\
%				Fourier components for\\
%				throttle and steering wheel angle}} &
\multirowcell{3}{$\approx$ 27.5 min} &
\multirowcell{1}{5} &
\multirowcell{1}{95\% }\\\cline{7-8}
&&&&&&
\multirowcell{1}{15} &
\multirowcell{1}{89\% }\\\cline{7-8}
&&&&&&
\multirowcell{1}{35} &
\multirowcell{1}{82\% }\\
\hline

\multirowcell{4}{Ezzini et al. 2018~\cite{Ezzini2018}} &
\multirowcell{1}{\shortstack{Controlled driving study~\cite{kwak2016}}} &
\multirowcell{4}{Extra Trees} &
\multirowcell{4}{\shortstack{N.A. / Variable}} &
\multirowcell{1}{\shortstack{\\51 CAN-bus signals}} &
%\multirowcell{2}{\shortstack{\\unclear, \\ not well described}} &
5 min & 10 & 99.92\% \\ \cline{2-2}\cline{5-8}
&\multirowcell{2}{Controlled driving study~\cite{Schneegass2013}}&&&
\multirowcell{2}{\shortstack{\\GPS, brightness,  acc.,\\ \& physiological signals}}&
1 min & \multirowcell{2}{6} & 99.99\% \\\cline{6-6}\cline{8-8}
&&&&& 2 min - 5min & & 100.0\% \\ \cline{2-2}\cline{5-8}
&\shortstack{Controlled driving study~\cite{romera2016}}&&&\shortstack{\\Smart phone GPS, gyroscope, \& Acc. sensor signals}
& \shortstack{N.A.} & 10 & 76\% \\ 
\hline

\multirowcell{1}{Jeong et al. 2018~\cite{Jeong2018}} &
\multirowcell{1}{\shortstack{\\Controlled driving study}} &
\multirowcell{1}{\shortstack{CNN, SVM, ELM}} &
\multirowcell{1}{\shortstack{0.25 Seconds}} &
\multirowcell{1}{\shortstack{CAN-bus}} &
%\multirowcell{1}{\shortstack{Event Separation\\ Acceleration \& Braking}} &
\multirowcell{1}{75s per driver} &
\multirowcell{1}{4} &
\multirowcell{1}{88\%}\\
\hline

\multirowcell{1}{\shortstack{Marchegiani et al. 2018~\cite{Marchegiani2018}}} &
\multirowcell{1}{\shortstack{Controlled driving study}} &
\multirowcell{1}{\shortstack{SVM \& GMM}} &
\multirowcell{1}{\shortstack{5 \& 15 Seconds}} &
\multirowcell{1}{\shortstack{CAN-bus: Throttle \& Brake}} &
%\multirowcell{1}{\shortstack{\\Cepstral coefficients and\\
%				their first and second derivatives}} &
\multirowcell{1}{\shortstack{$\approx$190 min}} &
\multirowcell{1}{\shortstack{4}} &
\multirowcell{1}{\shortstack{83\%}}\\
\hline

\multirowcell{1}{\shortstack{\\Rettore et al. 2018~\cite{Rettore2018}}} &
\multirowcell{1}{\shortstack{Controlled-, and driving study}} &
\multirowcell{1}{\shortstack{Extra Trees}} &
\multirowcell{1}{\shortstack{120 Seconds}} &
\multirowcell{1}{\shortstack{19 CAN-bus, Smartphone, \& Virtual sensor signals}} &
%\multirowcell{1}{\shortstack{PCA of the signals}} &
\multirowcell{1}{\shortstack{$\approx$2.8h per driver}} &
\multirowcell{1}{\shortstack{10}} &
\multirowcell{1}{\shortstack{98\%}}\\
\hline

\multirowcell{1}{\shortstack{\\Bernardi et al. 2018~\cite{Bernardi2018}}} &
\multirowcell{1}{\shortstack{Controlled driving study}} &
\multirowcell{1}{\shortstack{Multi-layer Perceptron}} &
\multirowcell{1}{\shortstack{1 Minute}} &
\multirowcell{1}{\shortstack{16 CAN-bus \& Smartphone Signals}} &
%\multirowcell{1}{\shortstack{Raw signals}} &
\multirowcell{1}{\shortstack{N.A.}} &
\multirowcell{1}{\shortstack{10}} &
\multirowcell{1}{\shortstack{95\%}}\\
\hline

\multirowcell{1}{\shortstack{\\Luo et al. 2018~\cite{Luo2018}}} &
\multirowcell{1}{\shortstack{Controlled driving study}} &
\multirowcell{1}{\shortstack{Random Forest}} &
\multirowcell{1}{\shortstack{N.A.}} &
\multirowcell{1}{\shortstack{300 CAN-bus signals}} &
%\multirowcell{1}{\shortstack{Statistical features for each sigal}} &
\multirowcell{1}{\shortstack{$\approx$120min}} &
\multirowcell{1}{\shortstack{4}} &
\multirowcell{1}{\shortstack{89\%}}\\
\hline

\end{tabular}
\caption{Detailed overview of the related work.}
\label{rel_work_tab}
\end{table}
\end{landscape}

\bibliographystyle{IEEEtran}
\bibliography{refs}

\newpage
\begin{IEEEbiography}[{\includegraphics[width=1in,height=1.25in,clip,keepaspectratio]{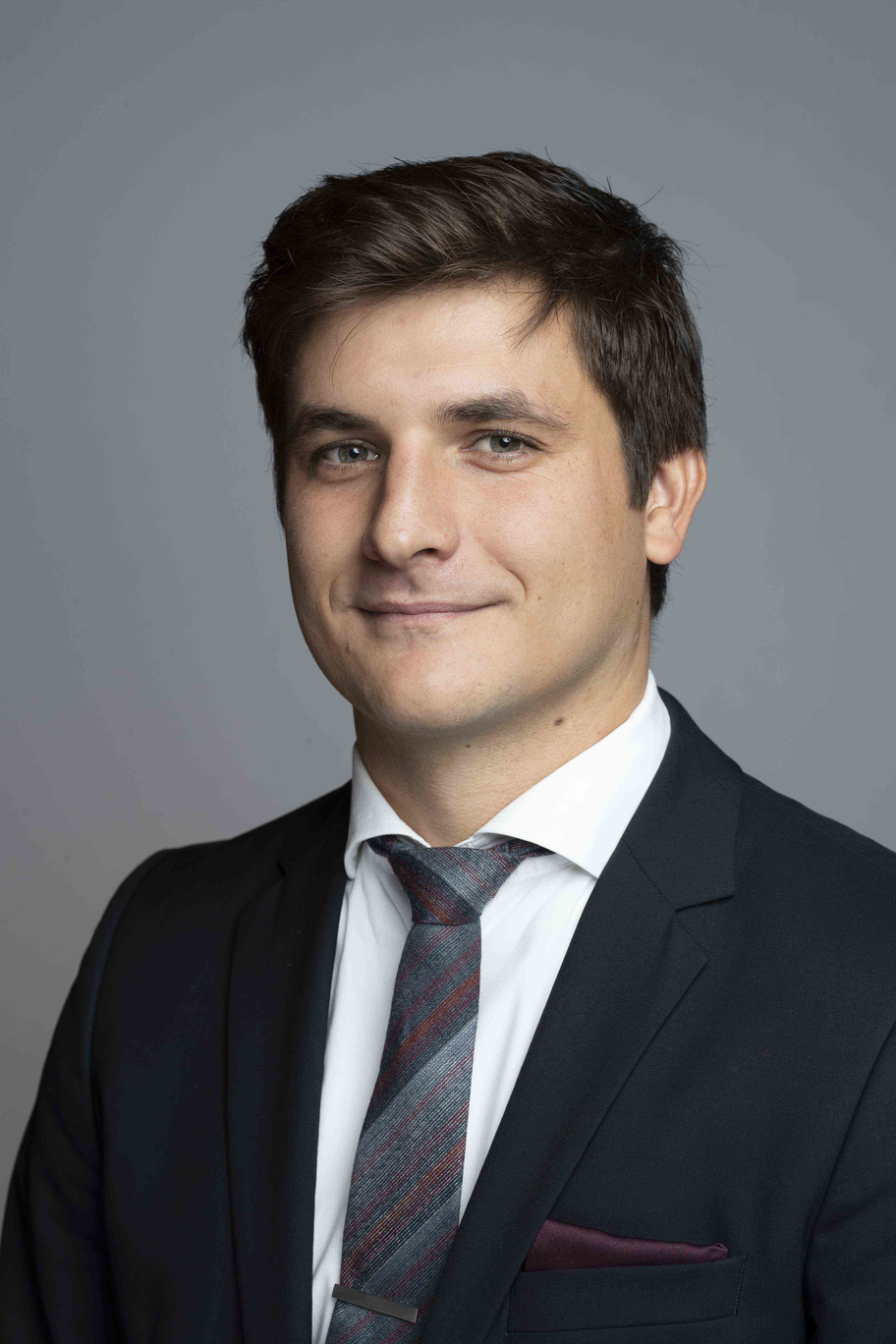}}]{Bernhard Gahr}
Bernhard Gahr is a PhD candidate at the Bosch IoT Lab at the University of St.Gallen (HSG) and investigates the potential of the connected car in the field of personalization.
He holds a diploma (M.Sc.) in electrical engineering and information technology from ETH Zurich. In addition, he completed an international stay for one year, studying at the LTH in Lund, Sweden.
Before joining the Bosch IoT lab, he worked for Open Systems on network security and for Bosch on combustion engine fuel injection systems. 
\end{IEEEbiography}
\begin{IEEEbiography}[{\includegraphics[width=1in,height=1.25in,clip,keepaspectratio]{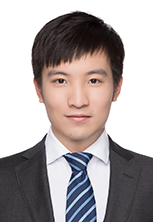}}]{Shu Liu}
Shu Liu is a PhD candidate at the Bosch IoT Lab at ETH Zurich. His research interest lies in the area of Machine Learning and Computer Vision in the context of Internet of Things / V2V and Autonomous Driving.
He holds a M.Sc in Electrical Engineering from ETH Zurich and a B.Sc from TU Munich. Before joining the Bosch IoT Lab he worked as a Research Assistant at the Computer Vision and Geometry Group of ETH Zurich in cooperation with MIT.
\end{IEEEbiography}
\begin{IEEEbiography}[{\includegraphics[width=1in,height=1.25in,clip,keepaspectratio]{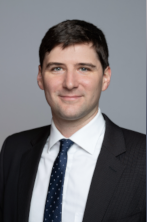}}]{Kevin Koch}
Kevin Koch is a PhD candidate at the Bosch IoT Lab at the University of St.Gallen (HSG). His research focuses on the potential of the connected car for driver state analysis and driving performance.
Kevin holds a Diploma (M.Sc.) in Information Systems from Technische Universit\"{a}t M\"{u}nchen (TUM) and a bachelor degree (B.Sc.) in Business Information Technology from Frankfurt School of Finance \& Management. In addition, he completed several international stays, including a research stay at the National Institute of Informatics in Tokyo (NII).
\end{IEEEbiography}
\begin{IEEEbiography}[{\includegraphics[width=1in,height=1.25in,clip,keepaspectratio]{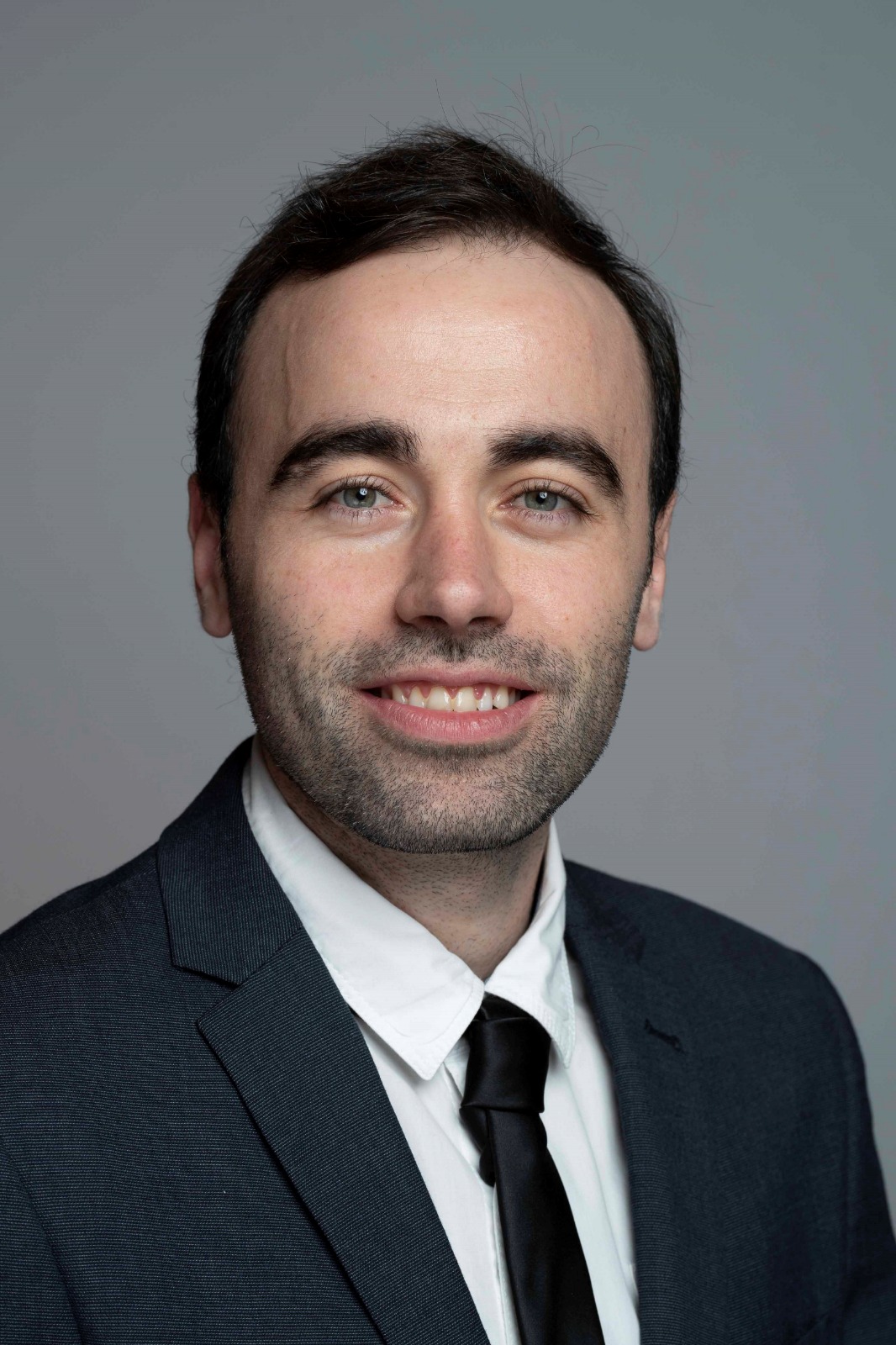}}]{Filipe Barata}
Filipe Barata is a Ph.D. candidate and doctoral researcher at the Center for Digital Health Interventions at ETH Zurich.  He employs artificial intelligence in the emerging field of patient-centered health information systems. In particular, he is investigating how to develop an effective, smartphone-based self-monitoring system for asthma.
He holds a diploma (M.Sc.) in electrical engineering and information technology from ETH Zurich. Before joining the Center for Health Interventions, he worked in the research department of Bruker Biospin, where he developed algorithms for the autonomous interpretation of NMR spectra. In his professional career he gained experience in several research laboratories, including the Department of Biomedical Data Science at the Geisel School of Medicine at Dartmouth College and the R\&D department of LMS International (Siemens).
\end{IEEEbiography}
\newpage
\begin{IEEEbiography}[{\includegraphics[width=1in,height=1.25in,clip,keepaspectratio]{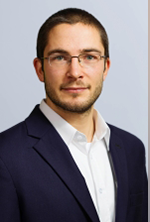}}]{Andr\'{e} Dahlinger}
Andr\'{e} Dahlinger received his doctoral degree from the University of St.Gallen (HSG) in 2018 and investigated the potential of the connected car for improving driving performance and car security.
He holds a Diploma (M.Sc.) in psychology and a B.Sc. in economics from the Philipps-University of Marburg. After spending one semester studying economics in Salamanca, Spain, he wrote his Diploma thesis in the field of behavioral economics (“The Effect of Ownership on Decisions under Risk”) and his Bachelor thesis in microeconomics (“Modeling Emotions in Microeconomics”).
\end{IEEEbiography}
\begin{IEEEbiography}[{\includegraphics[width=1in,height=1.25in,clip,keepaspectratio]{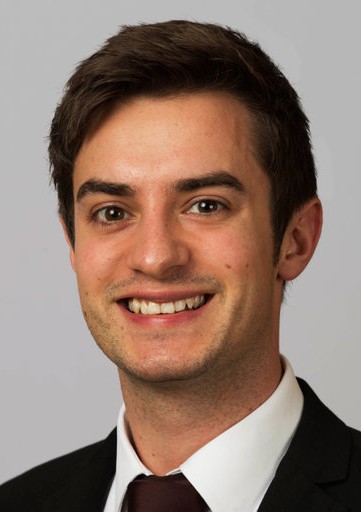}}]{Benjamin Ryder}
Benjamin Ryder received his doctoral degree from ETH Zurich in 2018 and investigated the potential of the connected car for improving driving performance and car security.
Ben graduated with an MEng in Computer Science from Imperial College London, United Kingdom, where his master’s thesis investigated the potential of the Microsoft Kinect as a touch screen input device. Benjamin undertook a research placement at the Hamyln Centre for Robotic Surgery, and after graduating worked for two years as a Technical Consultant with BAE Systems on a variety of projects focusing on autonomously detecting fraud for Insurance and Finance companies.
\end{IEEEbiography}
\begin{IEEEbiography}[{\includegraphics[width=1in,height=1.25in,clip,keepaspectratio]{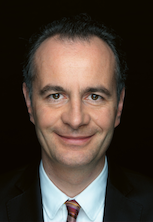}}]{Elgar Fleisch}
Elgar Fleisch, studied information systems at the University of Vienna (Austria). He received his doctoral degree from the University of Economics and Business Administration Vienna in 1993 for his work in artificial intelligence in production planning. Between 1994 and 2000 he worked as a postdoc in information management at the University of St. Gallen (Switzerland). In 2000 Elgar Fleisch was elected Associate Professor at the University of St. Gallen, becoming a full professor in 2002. In 2004 he was elected Professor for Information Management at ETH Zurich and started a double professorship with the University of St. Gallen.
\end{IEEEbiography}
\begin{IEEEbiography}[{\includegraphics[width=1in,height=1.25in,clip,keepaspectratio]{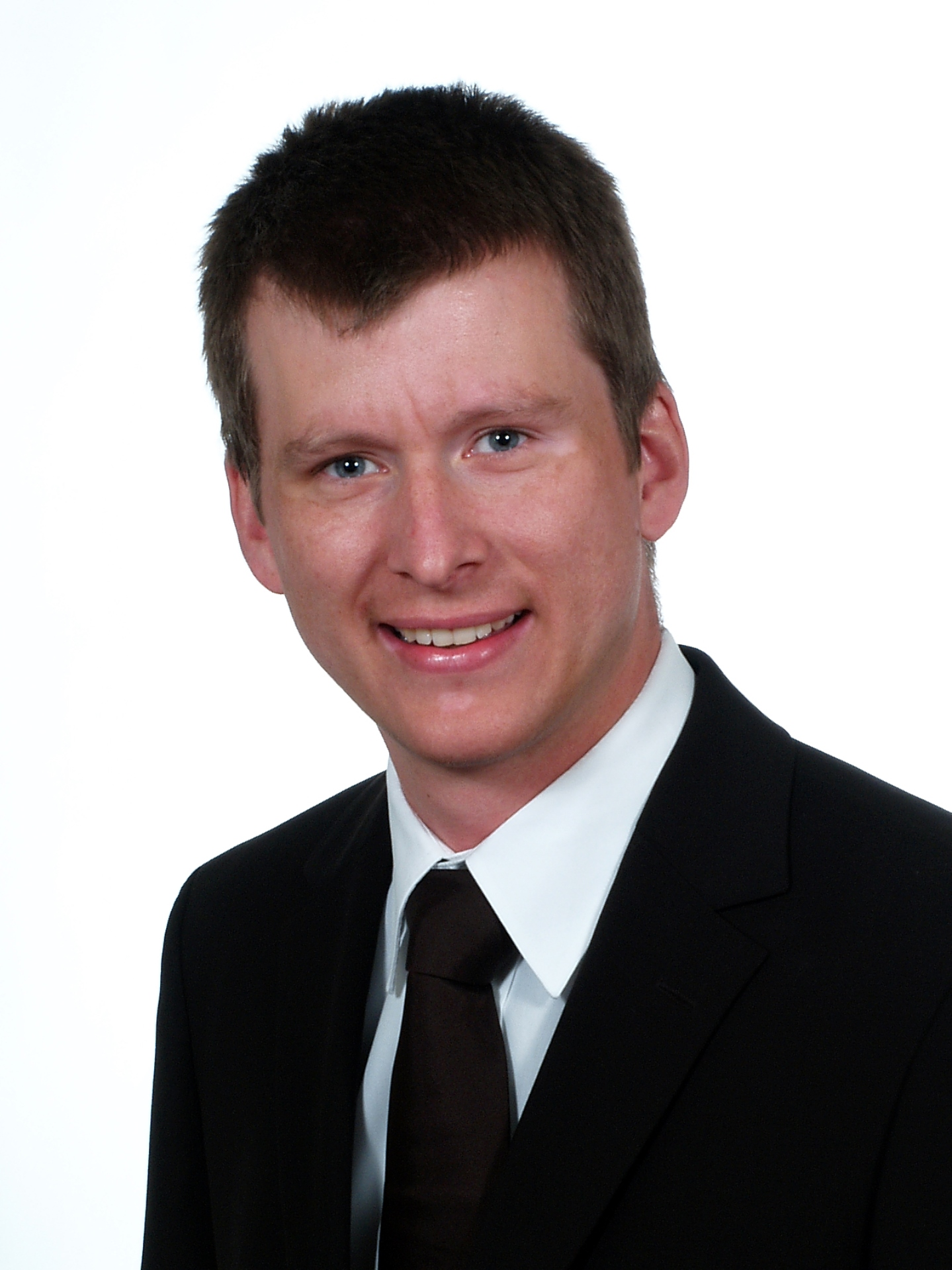}}]{Felix Wortmann}
Felix Wortmann is Assistant Professor of Technology Management at the University of St. Gallen. Furthermore, he is the Scientific Director of the Bosch IoT Lab, a research collaboration between Bosch, the University of St. Gallen and ETH Zurich. His research interests include the Internet of Things, blockchain, big data, and digital innovations in mobility, energy, and health. Felix Wortmann received a BScIS and MScIS from the University of Muenster, Germany, and a PhD in Management from the University of St. Gallen. He gained several years of industry experience in a German-based multinational software corporation.
\end{IEEEbiography}

\end{document}